# A CLIP-based Uncertainty Modal Modeling (UMM) Framework for Pedestrian Re-Identification in Autonomous Driving

**Jialin Li[1,4], Shuqi Wu[2,5], Ning Wang[3,6]**
[1] Jiangsu Zhenjiang Technician College, Jiangsu, China
[2] Amazon, New York, USA, 10001
[3] School of Computer Science, Robotics Institute, Carnegie Mellon University, Pittsburgh, USA

[4] 1812503968@qq.com

[5] shuqiwu2023@u.northwestern.edu

[6] ningwang@alumni.cmu.edu

**Abstract.** Re-Identification (ReID) is a critical technology in intelligent perception systems, especially within autonomous driving, where onboard cameras must identify pedestrians across views and time in real-time to support safe navigation and trajectory prediction. However, the presence of uncertain or missing input modalities—such as RGB, infrared, sketches, or textual descriptions—poses significant challenges to conventional ReID approaches. While large-scale pre-trained models offer strong multimodal semantic modeling capabilities, their computational overhead limits practical deployment in resource-constrained environments. To address these challenges, we propose a lightweight Uncertainty Modal Modeling (UMM) framework, which integrates a multimodal token mapper, synthetic modality augmentation strategy, and cross-modal cue interactive learner. Together, these components enable unified feature representation, mitigate the impact of missing modalities, and extract complementary information across different data types. Additionally, UMM leverages CLIP's vision-language alignment ability to fuse multimodal inputs efficiently without extensive fine-tuning. Experimental results demonstrate that UMM achieves strong robustness, generalization, and computational efficiency under uncertain modality conditions, offering a scalable and practical solution for pedestrian re-identification in autonomous driving scenarios.

## 1.Introduction

Person Re-Identification (ReID) is a core task in computer vision, aimed at recognizing individuals across different cameras and timeframes. In autonomous driving, ReID plays a critical role in pedestrian tracking, trajectory prediction, and collision avoidance. With the widespread adoption of vehicle-mounted cameras and multimodal sensors, ReID is evolving beyond traditional surveillance, requiring higher real-time performance, robustness to complex environments, and adaptability to diverse sensor modalities.

Traditional ReID approaches focused on single-modality RGB images in controlled environments, but they fail under challenges like motion blur, occlusion, or poor lighting commonly encountered in autonomous driving. Cross-modal ReID has extended this by incorporating infrared or sketch-based data, yet these methods are limited to fixed modality combinations and struggle with uncertain or missing input modalities—a frequent issue in real-world driving scenarios.

Recent developments in multimodal learning and pre-trained models such as CLIP have demonstrated promising results in bridging vision and language through contrastive learning. However, such models are often too resource-intensive for deployment on automotive platforms with limited computing capacity. Furthermore, high model complexity and missing sensor data hinder real-time application.

To address these challenges, this paper proposes a lightweight Uncertainty Modal Modeling (UMM) framework tailored for multimodal pedestrian ReID in autonomous driving. The framework integrates three key components: a unified multimodal token mapper for consistent representation across RGB, infrared, sketch, and text inputs; a synthetic augmentation strategy to simulate missing modalities; and a cross-modal interaction module that captures complementary information. Together, these components enhance robustness, scalability, and real-time applicability. The proposed UMM framework bridges the gap between theoretical advancements and practical deployment, offering a viable solution for multimodal pedestrian recognition in intelligent driving systems.

## 2.Literature Review

Single-Modality Person ReID focuses on retrieving visible images of specific pedestrians. The primary challenge in this domain lies in mitigating the negative effects on recognition performance caused by variations in camera viewpoints, background interference, posture changes and occlusion. Yi et al. [1] proposed a Siamese network designed to bring images of the same pedestrian from different views closer while pushing those of different individuals apart, thereby enabling the learning of view-invariant features. To address camera view and occlusion challenges, Dong et al. [2] introduced a multi-view information integration and propagation approach, which extracts robust and comprehensive features. Song et al. [3] tackled background noise by using pedestrian masks and integrating an attention module to suppress irrelevant background information.

Multimodal learning methods aim to leverage the complementary characteristics of various modalities to capture task-specific semantics [4]. Recently, multimodal transformers [5] have gained attention as unified models that combine different modality inputs through token concatenation, instead of independently learning modality-specific and cross-modality representations. Despite their potential, most multimodal approaches [6] rely on the assumption that all modalities are available during training or inference, an assumption that is often violated in practical scenarios. To address this limitation, researchers have developed methods [7] that can handle missing modalities. For instance, ImageBind [8] aligns features from various modalities into a shared feature space using contrastive learning, anchored to a base modality. SMIL [9] employs Bayesian Meta-Learning to estimate the latent features of absent modalities, while GCNet [10] incorporates graph neural networks to capture temporal and speaker dependencies, optimizing classification and reconstruction tasks simultaneously.

In real-world ReID applications, the uncertainty of the target modality poses significant challenges. For instance, depending on the environment and conditions, the available data could originate from diverse sources such as visible light images, infrared images, sketches, or even textual descriptions. This variability complicates the design of ReID systems, as traditional methods are often tailored to specific modality pairs and struggle to adapt when the input modality differs or is incomplete. Chen et al. [11] proposed a modality-agnostic retrieval framework that integrates RGB, sketch and text data, enabling learning and fusion of modality-specific features for both unimodal and multimodal tasks. This approach supports any combination of the three modalities, thereby increasing adaptability and reducing constraints. However, it does not incorporate IR data and involves a complex design with limited scalability. To address the challenge of diverse input modalities, this paper proposes a lightweight multimodal token mapper, which effectively transforms multimodal data into a unified feature space. This design not only facilitates seamless integration of information from different modalities but also reduces computational complexity, ensuring efficient and scalable performance in multimodal ReID tasks. In addition, this paper introduces a synthetic augmentation strategy that generates simulated features across multiple modalities to address the issue of missing modalities in real-world data, thereby enhancing the model's generalization ability by bridging the domain gap between training and real-world deployment. Finally, a cross-modal cue interactive learner is designed to capture complementary cues from cross-modal features, improving the model's performance in zero-shot learning scenarios.

## 3. METHODOLOGY

*3.1 Multi-Modal Data Representation*

Let the dataset consist of images and descriptions of N unique identities, where each identity i (i = 1,2, ..., N) is represented across four modalities:

RGB images (R): High-resolution color images capturing the visible spectrum, containing rich texture and appearance details.

IR images (I): Thermal images obtained in low-light or nighttime conditions, representing the person in heatmap-like forms.

Sketch images (S): Abstract representations of persons, often drawn manually or automatically generated, providing minimalistic structural information.

Text descriptions (T): Natural language descriptions of a person's attributes, such as clothing color, height and other semantic features.

For each person i, we define a multi-modal data tuple as:
$$D_i = \{x_i^R, x_i^S, x_i^I, x_i^T\} \tag{1}$$

where $x_i^M (M \in \{R, S, I, T\})$ denotes the data sample of person i in a specific modality.

*3.2 Cross-Modal Matching Objective*

Given a query q from modality $M_q$ and a gallery set $G = \{g_1, g_2, ..., g_M\}$ from modality $M_g$, the task is to learn a mapping function $f: X \rightarrow \mathbb{R}^d$, where d is the dimension of the shared embedding space. The mapping should minimize the inter-class similarity and maximize the intra-class similarity across all modalities. Ultimately, the similarity between samples of the same identity across same/different modalities should be greater than the similarity between samples of different identities within the same/different modality:

$$\text{Similarity}(f(q), f(g)) > \text{Similarity}\left(f(q), f(\hat{g})\right) \tag{2}$$

where $g, \hat{g} \in G$, g and q belong to the same identity, $\hat{g}$ and q do not belong to the same identity.

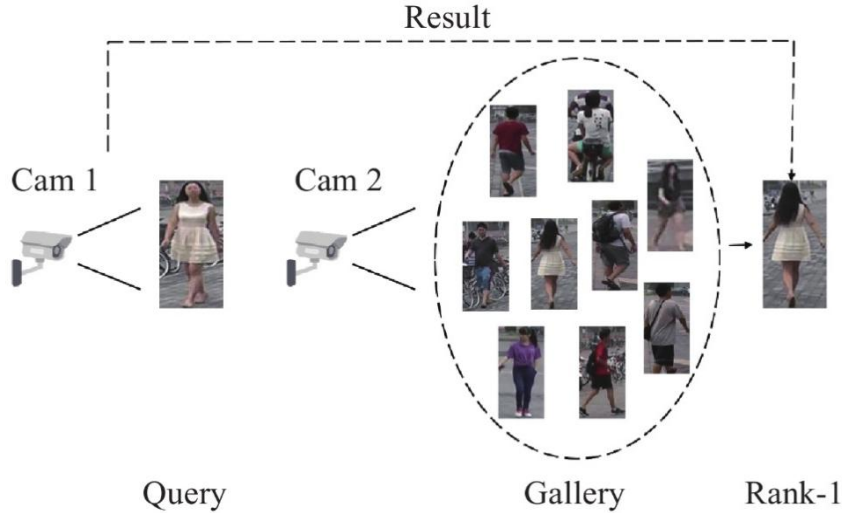

**Figure 1.** Pedestrian retrieval process diagram

### 3.3 Uncertainty Modal Modeling (UMM) Framework

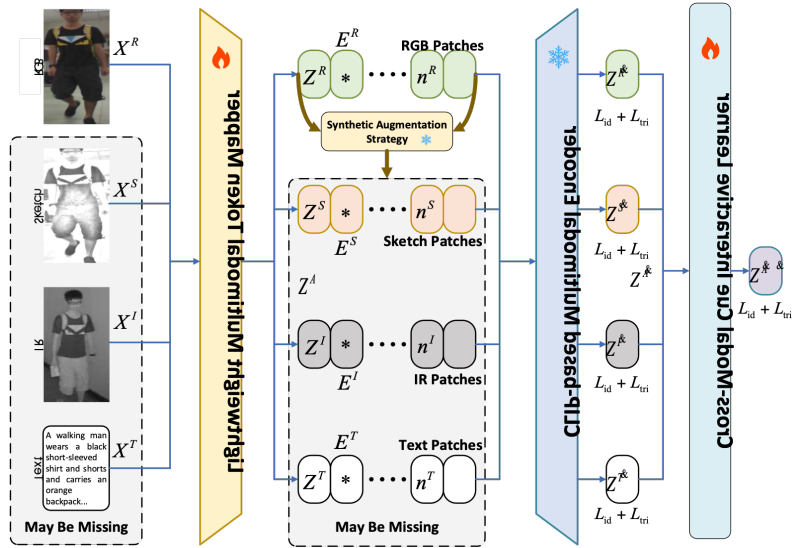

**Figure 2**. Framework diagram of Uncertainty Modal Modeling

To unify diverse modalities into a shared feature space for multimodal ReID, we introduce a Lightweight Multimodal Token Mapper (LMTM). This mapper processes input data from RGB, IR, sketch and text modalities, aligning them for further processing by a shared encoder.

Image Tokenization: For visual modalities (RGB, IR and sketch), channel alignment is achieved by replicating the single-channel IR and sketch images to match the three-channel RGB format. Unlike traditional tokenization methods that may introduce training instability, we adopt an IBN-style [12] tokenizer, as illustrated in Figure 3. This tokenizer integrates convolution, batch normalization and ReLU layers to enhance feature extraction and stabilize training.

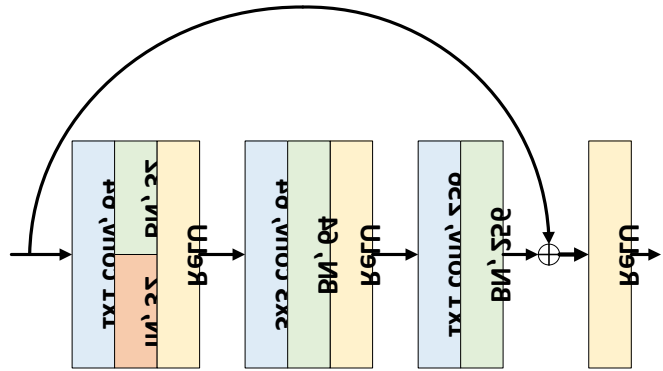

**Figure 3.** IBN-style Token Mapper

Traditional Batch Normalization (BN) relies on batch statistics for normalization, making it suitable for handling global features. However, when there are significant differences between modalities (such as RGB and IR images), BN is susceptible to interference from these modality differences. On the other hand, Instance Normalization (IN) normalizes based on statistics from individual samples, focusing more on local features and helping to preserve modality-specific fine-grained information. However, using IN alone can result in poorer distinguishability between instances, leading to the degradation of identity cues. Therefore, the IBN strategy combines these two normalization methods to enhance modality consistency while preserving identity discriminability, achieving an initial alignment of the three visual modalities. Specifically, BN handles cross-sample consistency and reduces modality differences, while IN strengthens sample-level normalization and aids in feature alignment. In addition to enhancing the adaptability of modality alignment, IBN significantly improves training stability, especially in small batch training.

Text Tokenization: Following prior works, we utilize the pre-trained CLIP encoder to map textual descriptions into a high-dimensional embedding space. Each word is converted into tokens, projected through embedding layers to create a sequence of word representations. Notably, the CLIP encoder is used in a frozen state, and its weights are not updated during training to preserve the rich semantic knowledge learned from large-scale pre-training.

Unified Multimodal Embedding: The embeddings from different modalities are concatenated to form a unified representation. A learnable token is appended to the sequence of embeddings, and positional encodings are added to enhance spatial information:

$$E^A = \{Z^A, E^R, E^S, E^I, E^T\} + E^{Pos} \tag{3}$$

$$E \in \mathbb{R}^{n \times D}, E^{Pos} \in \mathbb{R}^{(n+1) \times D} \tag{4}$$

where $Z^A$ is used to learn multimodal representation relationships, $E^{Pos}$ adds location information to multimodal embedding. This allows the model to process multimodal data seamlessly, enabling effective alignment across heterogeneous modalities. By integrating this multimodal tokenizer into our framework, we ensure a consistent and efficient transformation of diverse inputs, setting the stage for robust feature extraction and cross-modal alignment.

## 4. Experiments and Analysis

### 4.1 Dataset

The experiments utilize three publicly available datasets for training: SYNTH-PEDES for RGB-Text (R-T) pairs, LLCM for RGB-Infrared (R-I) images, and MaSk1K for RGB-Sketch (R-S) images. For evaluation, five benchmark datasets are selected to assess the zero-shot performance: Market1501 for RGB-to-RGB (R->R), SYSU-MM01 for Infrared-to-RGB (I->R), PKU-Sketch for Sketch-to-RGB (S->R), CUHK-PEDES for Text-to-RGB (T->R), and Tri-CUHK-PEDES for Text+Sketch-to-RGB (S+T->R). Detailed dataset statistics can be found in **Table 1**, with further information available in the respective original papers.

Table 1 The statistics of datasets used in experiments.

| Partition | Dataset | #ID | #RGB | #IR | #Sketch | #Text |
|---|---|---|---|---|---|---|
| Train | SYNTH-PEDES[39] | 312,321 | 4,791,771 | - | - | 12,138,157 |
| | LLCM[40] | 1,064 | 25,626 | 21,141 | - | - |
| | MaSk1K[41] | 996 | 32,668 | - | 4,763 | - |
| Test | Market1501[42] | 1,501 | 32,668 | - | - | - |
| | SYSU-MM01[43] | 491 | 30,071 | 15,792 | - | - |
| | PKU-Sketch[44] | 200 | 400 | - | 200 | 200 |
| | CUHK-PEDES[45] | 13,003 | 40,206 | - | - | 80,412 |
| | Tri-CUHK-PEDES[46] | 13,003 | 40,206 | - | 40,206 | 80,412 |

We adopt CLIP-ViT as the frozen backbone and employ a progressive training strategy. In the first 40 epochs, we train on 32 paired RGB-Text samples from SYNTH-PEDES with synthetic IR and Sketch embeddings, simulating missing modalities. In the following 80 epochs, training incorporates all datasets: SYNTH-PEDES uses the same sampling strategy, while LLCM and MaSk1K utilize only real RGB-IR and RGB-Sketch pairs. Multimodal embeddings are constructed using available modalities to enhance generalization under incomplete modality conditions.

### 4.2 Experimental results

To evaluate the performance of the proposed UMM, we conduct a comprehensive comparative analysis involving state-of-the-art cross-modal and multimodal ReID methods. These comparisons include unimodal generalized methods, cross-modal ReID approaches, multimodal techniques, and large-scale pre-trained ReID models, all within a zero-shot setting. As summarized in Table 2, existing large-scale pre-trained models, with the exception of PLIP, exhibit suboptimal performance in the zero-shot scenario. The UMM achieves competitive results on unimodal R->R retrieval tasks while consistently surpassing cross-modal ReID methods in all cross-modal tasks. Notably, the UMM demonstrates superior capability in integrating multiple modalities, addressing the limitations of existing methods that struggle to generalize to unseen modalities.

**Table 2** Zero-shot performance on cross-modal retrieval is reported in terms of the best Rank-1 and mAP metrics. Results marked with * indicate those reproduced by the authors.

| Type | Method | R->R | | I->R | | S->R | | T->R | |
|---|---|---|---|---|---|---|---|---|---|
| | | Rank-1 | mAP | Rank-1 | mAP | Rank-1 | mAP | Rank-1 | mAP |
| Pre-train | LuPerson-NL[48] | 24.6* | 11.6* | - | - | - | - | - | - |
| | PLIP[39] | 80.4 | 59.7 | - | - | - | - | 57.7 | - |
| | APTM[49] | 5.3* | 3.5* | - | - | - | - | 9.6* | 2.7* |
| Unimodal | OSNet-IBN[50] | 73.0 | 44.9 | - | - | - | - | - | - |
| | M3L[51] | 78.3 | 52.5 | | | | | | |
| | OSNet-AIN[52] | 73.3 | 45.8 | - | - | - | - | - | - |
| Cross-modal | AGW[53] | 17.3* | 6.9* | 18.2* | 19.1* | - | - | - | - |
| | IRRA[54] | 66.6* | 40.5* | - | - | - | - | 30.1* | 25.3* |
| | UNIReID[46] | 19.0* | 8.2* | - | - | 69.8 | 73.0 | 11.6* | 9.7* |
| Multimodal | UMM (Ours) | 78.6 | 58.4 | 56.4 | 50.7 | 70.1 | 73.2 | 52.4 | 42.1 |

In the cross-modal testing phase, the Synthetic Augmentation Strategy is not used to generate embeddings for all modalities. This decision is made to avoid the introduction of excessive noise, as only 1/4 of the embeddings would be real, with the remaining 3/4 being synthetic, thereby increasing inference cost and degrading performance. For instance, in the I->task, the query infrared image is processed through the UMM to extract $Z^I$, representing the infrared embedding. For the candidate RGB images, the Synthetic Augmentation Strategy is employed to generate synthetic infrared embeddings $Z^{\dot{I}}$. $Z^{\dot{R}}$ and $Z^{\dot{I}}$ are then fused with the original RGB embeddings using the Cross-Modal Cue Interactive Learner. By leveraging this approach, the disparity between the query and candidate modalities is significantly alleviated, resulting in enhanced retrieval accuracy.

*4.3 Visualization*

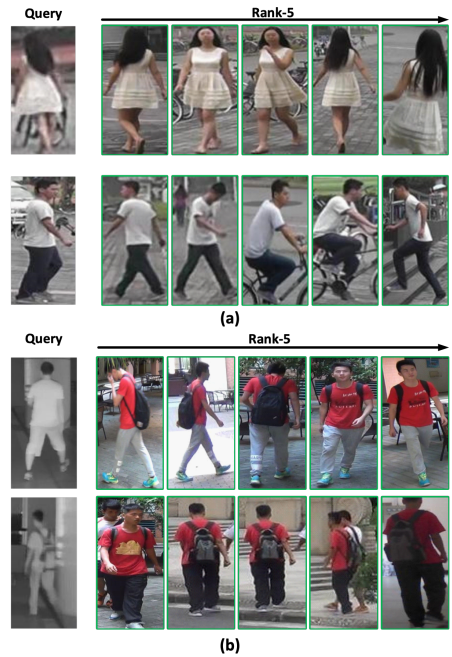

**Figure 4.** (a) Visualization of R->R partial search results; (b) Visualization of I->R partial search results.

From R→R partial search results, the proposed UMM approach demonstrated promising results in generalization wrt change of viewpoints and pedestrian poses. In the first row, UMM was able to successfully retrieve image patches of the lady in white dress invariant of viewpoints when the query is captured from the back. Similar story in the second example, where the man riding bicycle is successfully retrieved with the query of him walking.

For I→R partial search, the accuracy relies more on the richness of the visual features in the IR image query. This is expected as IR images is considered a much sparser representation as compared to RGB images due to missing color information and crisp object details as radiation signals usually generate blurry edges. In the first example, where the query target pedestrian is in clothing of plain colors, UMM failed to retrieve the correct target agent. In the second example, the backpack on the target agent serves as additional visual features and helps UMM in retrieving correct image patches.

Both S→R & T→R results shares similar characteristics as R→R results and show high retrieval accuracy that are invariant to viewpoints and lighting conditions.

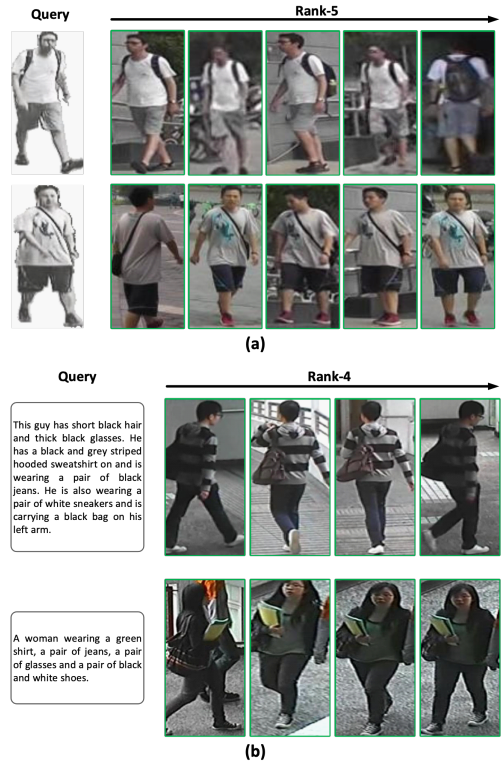

**Figure 5.** (a) Visualization of S->R partial search results; (b) Visualization of T->R partial search results

## 5.Conclusion

This study proposes a lightweight Uncertainty Modal Modeling (UMM) framework to address key challenges in pedestrian Re-Identification (ReID) for autonomous driving, including modality uncertainty, data incompleteness, and cross-modal fusion. Designed for vehicle-mounted camera systems, UMM enhances the robustness of person recognition across varying road conditions. It integrates three core modules: a multimodal token mapper for unified representation, a synthetic augmentation strategy to compensate for missing inputs, and a cross-modal cue interactive learner to capture complementary features. The framework efficiently fuses RGB, infrared, sketch, and text modalities while maintaining high performance under incomplete data conditions. Leveraging CLIP's vision-language alignment, UMM achieves strong zero-shot generalization without extensive fine-tuning, making it ideal for deployment on resource-limited automotive platforms. Experiments show UMM outperforms existing methods in generalization, efficiency, and accuracy. Additionally, it includes built-in anonymization and encryption to ensure ethical and secure use of sensory data, offering a practical and responsible solution for real-world autonomous systems.

In conclusion, this work focuses on the challenges of pedestrian ReID in autonomous driving space due to factors such as rapid change of viewpoints, complex lighting and motion blurs, etc., and proposes a novel framework for robust and efficient ReID through cross-modality fusion and synthetic augmentations. This approach demonstrates compelling performance on ReID retrieve accuracy and robustness towards viewpoints, lighting conditions, and pedestrian actions. By leveraging pre-trained CLIP model for feature alignment, the proposed approach was able to achieve promising zero-shot accuracy without extensive training, which opens up more inspirations on approaches to utilize and ground large pre-trained foundation models for specialized applications.